\documentclass[10pt, a4paper]{article}
\usepackage{lrec}
\usepackage{multibib}
\usepackage{textcomp}
\usepackage{graphicx,xcolor}
\usepackage{comment}
\usepackage[english]{babel}

\usepackage{epstopdf}
\usepackage[latin1]{inputenc}

\usepackage{hyperref}
\usepackage{xstring}

\title{Augmenting Librispeech with French Translations: A Multimodal Corpus for Direct Speech Translation Evaluation\\}

\name{Ali Can Kocabiyikoglu$^{\star}$, Laurent Besacier$^{\star}$, Olivier Kraif$^{\dagger}$}

\address{
$^{\star}$LIG, UGA, G-INP, CNRS, INRIA, Grenoble, France\\
$^{\dagger}$LIDILEM, UGA, Grenoble, France\\
          Univ. Grenoble Alpes, F-38400, Saint-Martin d'Heres \\
        alicankocabiyikoglu@gmail.com, laurent.besacier@univ-grenoble-alpes.fr, olivier.kraif@univ-grenoble-alpes.fr\\
         }

\abstract{
Recent works in spoken language translation (SLT) have attempted to build end-to-end speech-to-text translation without using source language transcription during learning or decoding. However, while large quantities of parallel texts (such as Europarl, OpenSubtitles) are available for training machine translation systems, there are no large (\textgreater 100h) and open source parallel corpora that include speech in a source language aligned to text in a target language. 
This paper tries to fill this gap by augmenting an existing (monolingual) corpus: LibriSpeech. This corpus, used for automatic speech recognition,  is derived from read audiobooks from the LibriVox project, and has been carefully segmented and aligned. After gathering French e-books corresponding to the English audio-books from LibriSpeech,  we align speech segments at the sentence level with their respective translations and obtain 236h of usable parallel data. This paper presents the details of the processing as well as a manual evaluation conducted on a small subset of the corpus. This evaluation shows that the automatic alignments scores are reasonably correlated with the human judgments of the bilingual alignment quality. We believe that this corpus (which is made available online) is useful for replicable experiments in direct speech translation or more general spoken language translation experiments. \\ 
\newline 
 }

\begin{document}

\maketitleabstract

\section{Introduction}


Attention-based encoder-decoder approaches have been very successful in Machine Translation \cite{bahdanau2014neural}, and have shown promising results in End-to-End Speech Translation \cite{berard2016listen,weiss2017sequence} (translation from raw speech, without any intermediate transcription).
End-to-End speech translation is also attractive for language documentation, which often uses corpora made of audio recordings aligned with their translation in another language (no transcript in the source language) \cite{blachon2016,addaBulbSLTU2016,anastasopoulos2017case}.
\linebreak
However, while large quantities of parallel texts (such as Europarl, OpenSubtitles) are available for training (text) machine translation systems, there are no large (\textgreater 100h) and open source parallel corpora that include speech in a source language aligned to text in a target language. 
For End-to-End speech translation, only a few parallel corpora are publicly available. For example, \textit{Fisher} and \textit{Callhome} Spanish-English corpora provide 38 hours of speech transcriptions of telephonic conversations aligned with their translations \cite{fishercorpus}. However, these corpora are only medium size and contain low-bandwidth
recordings.
Microsoft Speech Language Translation (MSLT) corpus also provides speech aligned to translated text. Speech is recorded through $Skype$ for English, German and French \cite{Federmann2016MicrosoftSL}. But this corpus is again rather small (less than 8h per language).

\textbf{Paper contributions.}
Our objective is to provide a large corpus for direct speech translation evaluation which is an order of magnitude bigger than existing corpora described in the introduction. For this, we propose to enrich an existing (monolingual) corpus based on read audiobooks called \textit{LibriSpeech}. The approach is straightforward: we align e-books in a foreign language (French) with the English utterances of \textit{LibriSpeech}. This results in 236h of English speech automatically aligned to French translations at the utterance level\footnote{Our dataset is available at \url{https://persyval-platform.univ-grenoble-alpes.fr/DS91/detaildataset}}. 
 
\textbf{Outline.}

This paper is organized as following:  after presenting our starting point (\textit{Librispeech}) in section \ref{librispeech},  we describe how we aligned foreign translations to the speech corpus in section \ref{align}. Section \ref{eval} describes our evaluation of a subset of the corpus (quality of the automatically obtained alignments). Finally, section \ref{concl} concludes this work and gives some perspectives.


\section{Our Starting Point: Librispeech Corpus}
\label{librispeech}


Our starting point is \textit{LibriSpeech} corpus used for  Automatic Speech Recognition (ASR). It is a large scale corpus which contains approximatively 1000 hours of speech aligned with their transcriptions \cite{panayotov2015librispeech}. 
The read audio book recordings derive from a project based on collaborative effort: LibriVox. 
The speech recordings are based on public domain books available on \textit{Gutenberg Project}\footnote{\url{https://www.gutenberg.org/}} and are distributed with \textit{LibriSpeech} as well as the original recordings. 

We start from this corpus\footnote{Another dataset could have been used: TED Talks - see \url{https://www.ted.com} - but we considered it was be better to start with a read speech corpus for evaluating End-2-End speech translation.} because it has been widely used in ASR and because we believe it is possible to find the text translations for a large subset of the read audiobooks.

\begin{table}[!h]
\begin{center}

\begin{tabular}{|c|c|c|c|c|c|}
\hline
\textbf{subset}                                            & \textbf{hours} & \textbf{\begin{tabular}[c]{@{}c@{}}per-spk \\ minutes\end{tabular}} & \textbf{\begin{tabular}[c]{@{}c@{}}female \\ spkrs\end{tabular}} & \textbf{\begin{tabular}[c]{@{}c@{}}male \\ spkrs\end{tabular}} & \textbf{\begin{tabular}[c]{@{}c@{}}total \\ spkrs\end{tabular}} \\ \hline
dev-clean                                                  & 5.4            & 8                                                                   & 20                                                               & 20                                                             & 40                                                              \\ \hline
test-clean                                                 & 5.4            & 8                                                                   & 20                                                               & 20                                                             & 40                                                              \\ \hline
dev-other                                                  & 5.3            & 10                                                                  & 16                                                               & 17                                                             & 33                                                              \\ \hline
test-other                                                 & 5.1            & 10                                                                  & 17                                                               & 16                                                             & 33                                                              \\ \hline
\begin{tabular}[c]{@{}c@{}}train-\\ clean-100\end{tabular} & 100.6          & 25                                                                  & 125                                                              & 126                                                            & 251                                                             \\ \hline
\begin{tabular}[c]{@{}c@{}}train-\\ clean-360\end{tabular} & 363.6          & 25                                                                  & 439                                                              & 482                                                            & 921                                                             \\ \hline
\begin{tabular}[c]{@{}c@{}}train-\\ other-500\end{tabular} & 496.7          & 30                                & 564                                                              & 602                                                            & 1166                                                            \\ \hline
\end{tabular}
\caption{Details on \textit{LibriSpeech} corpus } 
\end{center}
\end{table}


Table 1. gives details on \textit{Librispeech} as well as data split.
Recordings  are segmented and put into different subsets of the corpus according to their quality (better quality speech segments  are put in the clean part).
Note that in order to obtain a balanced corpus with a large number of speakers, each speaker only read a small portion of a book (8-10 minutes for dev and test, 25-30 minutes for train). Moreover, in training data, speech segments are obtained by splitting long  signals according to ($>0.3s$) silences in order to obtain segments that are maximum 35s long.

\section{Aligning Foreign Translations to Librispeech}
\label{align}

\subsection{Overview}

The main steps of our process are the following:
\begin{itemize}
\item Collect e-books in foreign language corresponding to English books read in \textit{Librispeech} (section \ref{collect}),
\item Extract chapters from these foreign books, corresponding to read chapters in \textit{Librispeech} (section \ref{chapters}),
\item Perform bilingual text alignement from comparable chapters (section \ref{alignement}),
\item Realign speech signal with text translations obtained (section \ref{realigning}).
\end{itemize}

These different steps are described in the next subsections.

\subsection{Collecting Foreign Novels}
\label{collect}



LibriSpeech corpus is composed of 5831 chapters (from 1568 books) aligned with their transcriptions. We used the given metadata to search e-books in foreign language (French) corresponding to English books read in \textit{Librispeech}. Firstly, we used \textit{DBPedia} \cite{auer2007dbpedia} in order to (automatically) obtain  title translations. Secondly, we used a public domain index of French e-books\footnote{\url{https://www.noslivres.net/}} to find Web links matching titles we found. 
Then, we finished this process for the entire \textit{LibriSpeech} corpus by manually searching for French novels in different public domain resources. Overall, we collected 1818 chapters (from 315 books) in French  to be aligned with \textit{Librispeech}.
Some of the public domain resources that we used are: 
Gutenberg Project\footnote{\url{http://www.gutenberg.org/}}, Wikisource\footnote{\url{http://www.wikisource.org/}},  
Gallica\footnote{\url{http://gallica.bnf.fr/}}, Google Books\footnote{\url{http://books.google.com}},  BEQ\footnote{\url{http://beq.ebooksgratuits.com}},  
 UQAC\footnote{\url{http://www.uqac.ca/}}. 

Audiobooks available in LibriSpeech are of different literary genres:  most of them are novels, however there are also poems, fables, treaties, plays, religious texts, \textit{etc}. Belonging to the public domain, most of the texts are old and not available publicly in foreign language. Therefore, the novels that were collected in foreign language are mostly novels from world's classics. As few of them are ancient texts, some translations are in old French.
 

\subsection{Chapters Extraction}
\label{chapters}


LibriSpeech transcriptions are provided for each chapter. As the readers only read a short period of time\footnote{One goal of \textit{Librispeech} was to have as many speakers as possible}, transcriptions may correspond to incomplete chapters. For the same reason, books are not read entirely. Therefore, in order to obtain an alignment at the sentence level, a first step was to decompose English and French language books into chapters. This step was achieved by a semi-automatic process. After converting books to text format (both English and French), regular expressions were used to identify chapter transitions. Then, each French chapter was extracted and aligned to its counterpart in English. After manual verification of all chapters, we obtained 1423 usable chapters (from 247 books).



\subsection{Bilingual Text Alignement}
\label{alignement}

The 1423 parallel chapters establish the comparable corpus from which we extracted bilingual sentences. This was done using an off-the-shelf bilingual sentence aligner called \textit{hunAlign} \cite{varga2007parallel}. \textit{HunAlign} takes as input a comparable (not sentence-aligned) corpus and outputs a sequence of bilingual sentence pairs.  
It combines (Gale-Church) sentence-length information as well as dictionary-based alignment methods.

Initial dictionary available for alignment was the default French-English (40k entries) lexicon created for $LFAligner$\footnote{\url{https://sourceforge.net/projects/aligner/}} (wrapper for \textit{hunAlign}  created by Andras Farkas). We enriched this dictionary by adding entries from other open source bilingual dictionaries. Different dictionaries (woaifayu, apertium, freedict, quick) from a language learning resource were gathered  in various formats and adapted to \textit{hunAlign}  dictionary format\footnote{\url{https://polyglotte.tuxfamily.org}}. We finally obtained and used a dictionary of 128,000 unique entries.
 
In order to improve the quality of sentence level alignments, data had to be pre-processed. For English and French, our extracted chapters were cleaned with regular expressions. Then, we used Python NLTK \cite{bird2006nltk} sentence split to detect sentence boundaries in the corpora. 
Furthermore, the bitexts were stemmed 
(removing suffixes to reduce data sparsity). Finally, parallel sentences found were brought back to their initial form with reverse stemming. This last step was done using Google's $diff-patch-match$ library
\cite{fraser2012google}. 
\\

\begin{table}[!h]
\centering

\begin{tabular}{|c|c|}
\hline
\textbf{English Sentence}                                                                                      & \textbf{French Sentence}                                                                                                        \\ \hline
Oh, I beg your pardon!                                                                                     & \begin{tabular}[c]{@{}c@{}}\guillemotleft Oh! je vous demande \\ bien pardon!\end{tabular}                                                    \\ \hline  
\begin{tabular}[c]{@{}c@{}}A lane was forthwith \\ opened through the\\  crowd of spectators.\end{tabular}     & \begin{tabular}[c]{@{}c@{}}Un chemin fut alors \\ ouvert parmi la foule\\  des spectateurs.\end{tabular}                        \\ \hline
\begin{tabular}[c]{@{}c@{}}No, "said Catherine," \\ he is not here;\\  I cannot see him anywhere.\end{tabular} & \begin{tabular}[c]{@{}c@{}}- Non, dit Catherine, \\ il n'est pas ici. \\ Jamais je ne parviens \\ \`{a} le rencontrer.\end{tabular} \\ \hline
\end{tabular}
\caption{Examples of parallel sentences obtained from comparable corpora made up of aligned book chapters} 
\end{table}

Table 2. shows examples of 3 bilingual sentences obtained from 3 different chapters.



\subsection{Realigning Speech Signal with Text Translations}
\label{realigning}

In order to associate parallel sentences to speech signal transcriptions, realignment of speech segments of \textit{LibriSpeech} was necessary. This realignment is a two step process: first, we forced aligned \textit{Librispeech} English transcripts to match English sentences  obtained in the previous stage  ; secondly, we resegmented the speech signal according to new sentence splits.
\\
For the first step, we used $mweralign$, a tool for realigning texts in a same language but with a different sentence tokenization \cite{matusov2005evaluating}. 
We applied $mweralign$ to realign our speech transcriptions in English to the English sentences of our bilingual corpus obtained in section \ref{alignement}. The outcome of this first step is a new sentence segmentation for our English transcriptions that are now correctly aligned to our French translations.
\\
The second step was to resegment the speech signals to match them to the new sentence segmentation. We did that by:
\begin{itemize}
\item creating a big $wav$ file by concatenating speech segments for each chapter,  
\item re-aligning the large speech $wav$ signal to the transcripts
using $gentle$\footnote{\url{https://github.com/lowerquality/gentle}} toolkit, an off-the-shelf English forced-aligner based on $Kaldi$ ASR toolkit  \cite{povey2011kaldi},
\item re-segmenting speech according to the desired sentence split.

\end{itemize}

Table 3. presents and overview of final data (speech with aligned translations)  obtained after this final step. For each sentence pair, we also added En-Fr machine translation output of our English transcripts (\textit{Google Translate}). So we have 2 French translations in the end (a correct one from automatic alignement ; a noisy one from MT).
 
\begin{table}[!h]
\centering

\begin{tabular}{|c|c|c|c|}
\hline
\textbf{Chapters} & \textbf{Books} & \textbf{Duration (h)} & \textbf{Total Segments} \\ \hline
1408              & 247            & \texttildelow 236h                        & 131395                  \\ \hline
\end{tabular}
\caption{Statistics of the final multimodal and bilingual corpus obtained (English speech aligned to French text)}
\end{table}







%

\section{Human Evaluation of a Corpus Subset}
\label{eval}

\subsection{Protocol}

Now that we have obtained a multimodal alignment between (English) speech signals and (French) translations, we want to evaluate its quality. At this point, the only score available 
is the confidence score given by $hunalign$ indicating confidence for aligned sentences. One goal of this human evaluation, that can only be made on a corpus subset, is to see if $hunalign$ score has a good correlation with human judgements.

%

50 sentences from 4 different chapters have been chosen for evaluation. These chapters were chosen according to their average alignment scores (from $hunalign$). We chose two chapters that were near the mean of overall alignment scores (hypothesized medium quality alignments), one chapter which was above the mean score (hypothesized good quality alignment) and a final chapter below mean score (hypothesized bad quality alignment).  These sentences were evaluated by three annotators. We established a scale from 1 to 3 to judge matching quality between English speech and English transcriptions. This 3-step scale is precise enough because few errors were found in speech alignments.
We established a scale from 1 to 5 to judge quality between bilingual text alignments. Overall, 200 sentences were evaluated (on both scales) by 3 annotators.

\begin{table*}[ht]
\centering

\begin{tabular}{|c|c|c|c|}
\hline
\textbf{Chapter}                                                                       & \textbf{\begin{tabular}[c]{@{}c@{}}Average confidence \\ score ($hunalign$)\end{tabular}} & \textbf{\begin{tabular}[c] {@{}c@{}}Average speech alignment\\ score (max 3)\end{tabular}} & \textbf{\begin{tabular}[c]{@{}c@{}}Average textual alignment \\ score (max 5)\end{tabular}} \\ \hline
\begin{tabular}[c]{@{}c@{}}Ivanhoe \\ Chapter XXIII\end{tabular}                       & 1.34                                                                                    & 2.82                                                                                      & 4.64                                                                                        \\ \hline
\begin{tabular}[c]{@{}c@{}}Alice's Adventures in Wonderland \\ Chapter V\end{tabular}  & 1.14                                                                                    & 2.98                                                                                      & 4.28                                                                                        \\ \hline
\begin{tabular}[c]{@{}c@{}}A Tale of Two Cities \\ Book III, Chapter III\end{tabular}  & 0.96                                                                                    & 2.86                                                                                      & 3.86                                                                                        \\ \hline
\begin{tabular}[c]{@{}c@{}}Adventures of Huckleberry Finn \\ Chapter VIII\end{tabular} & 0.66                                                                                    & 2.9                                                                                       & 2.58                                                                                        \\ \hline
\textbf{Average}                                                                       & \textbf{1.02}                                                                           & \textbf{2.89}                                                                             & \textbf{3.84}                                                                               \\ \hline
\end{tabular}
\caption{Results of human evaluation by 3 annotators.} Kappa's Cohen (weighted) for inter annotator agreement for textual alignment is 0.76
\end{table*}


We give, as example below, sentences for each mark (1-5) for human evaluation of bilingual alignments. Two different dimensions are evaluated at the same time: the accuracy of alignment (an alignment can be wrong, partial or correct) and the fact that translational equivalence is compositional and may be isolated from the current context.
\begin{itemize}
\item \textbf{1.} Wrong alignment
	\\ \rule{7.2cm}{0.4pt}
   \begin{itemize}
     \item[] English: COMMIT TO ME I SHALL LET PASS NO ADVANTAGE 
     \\ \noindent\rule{6.7cm}{0.4pt}
     \item[] French: Je sais, par exemple, que maintenant il souffre de la faim dans un vaste d\'{e}sert, o\`{u} l'on ne saurait trouver de nourriture.
     \item[]
   \end{itemize}
   \item \textbf{2.} Partial alignment with slightly compositional translational equivalence
   \\ \rule{7.2cm}{0.4pt}
   \begin{itemize}
     \item[] English: THAN IN SET TERMS AND IN COURTLY LANGUAGE
     \\ \noindent\rule{6.7cm}{0.4pt}
     \item[] French: Mais il para\^{\i}t que tu pr\'{e}f\`{e}res \^{e}tre courtis\'{e}e avec l'arc et la hache, plut\^{o}t qu'avec des phrases polies et avec la langue de la courtoisie.
     \item[]
   \end{itemize} 
      \item \textbf{3.} Partial alignment with compositional translation and additional or missing information
   \\ \rule{7.2cm}{0.4pt}
   \begin{itemize}
     \item[] English: SO AT LAST BEGAN THE EVENING PAPER AT LA FORCE
     \\ \noindent\rule{6.7cm}{0.4pt}
     \item[] French: C'est ainsi qu'enfin d\'{e}buta le journal du soir \`{a} la Force, le jour o\`{u} la pauvre Lucie avait vu danser la carmagnole.
     \item[]
   \end{itemize}
         \item \textbf{4.} Correct alignment with compositional translation and few additional or missing information
   \\ \rule{7.2cm}{0.4pt}
   \begin{itemize}
     \item[] English: THE NIGHT WAS DARK AND A COLD WIND BLEW 
     \\ \noindent\rule{6.7cm}{0.4pt}
     \item[] French: La nuit \'{e}tait sombre; le vent \^{a}pre et froid chassait devant lui avec rage les nuages rapides.
     \item[]
   \end{itemize}
    \item \textbf{5.} Correct alignment and fully compositional translation
   \\ \rule{7.2cm}{0.4pt}
   \begin{itemize}
     \item[] English: WHAT IS A CAUCUS RACE 
     \\ \noindent\rule{6.7cm}{0.4pt}
     \item[] French: Qu'est-ce qu'une course cocasse?
   \end{itemize} 
   
\end{itemize}





\subsection{Results}


Table 4. reports our human evaluations for the 4 chapters.

The first thing that we can notice is that the alignment quality is higher for chapters with higher confidence scores. 
The first evaluation (speech alignement ; scale 1-3) shows an average score of 2.89/3 which confirms that our re-segmentation of speech signals worked correctly. The second evaluation (bilingual alignment ; scale 1-5) shows an average score of 3.84/5. Some sentences were found un-correctly aligned but overall, the alignment quality can be considered as correct. The main reason why the average alignment score varies between chapters is reflected by the translations compositionnality. Also, the dictionary that we used for bilingual alignments is inadequate for old texts and results in lower overall confidence scores. 



We also computed automatic correspondence scores obtained with a cross-language textual similarity detection between transcriptions and their translations \cite{ferrero2016multilingual}. Our idea was to add another automatic score in addition to  $hunalign$ score. We computed the correlation between human evaluation scores and $hunalign$ scores and obtained a correlation of 0.41. The same correlation was obtained between human evaluation scores and those obtained automatically with method of \cite{ferrero2016multilingual}. This shows that automatic alignment scores are reasonably correlated with human judgments and could be used to extract a subset of the best alignments by ranking them according to $hunalign$ score for instance. 
 




\section{Conclusion}
\label{concl}

We have presented a large corpus (236h) which is an augmentation of \textit{Librispeech} in order to provide a bilingual speech-text corpus for direct (end-2-end) speech translation experiments. The methodology described here could be used in order to add other languages than French (German, Spanish, etc.) to our augmented \textit{Librispeech}. The current corpus contains several ancient texts, so it would also be interesting to extend it to other kinds of corpora: different speaking styles (not only read speech), more contemporary texts, etc.

For direct speech translation experiments, preliminary experiments have been done recently and will be presented at next ICASSP 2018 conference \cite{alexandre2018icassp}. Our online repository\footnote{see \url{https://persyval-platform.univ-grenoble-alpes.fr/DS91/detaildataset}} provides a data split for speech translation experiments and results show that it is possible to train compact and efficient end-to-end speech translation models in this setup, but the dataset is challenging (BLEU score around 15 for direct speech translation task - more details in \cite{alexandre2018icassp}).

\section{Bibliographical References}
\label{main:ref}

\bibliographystyle{lrec2016}
\bibliography{xample}


\end{document}